\documentclass[]{clv2}

\title{A Context-theoretic Framework for Compositionality in Distributional Semantics}
\runningtitle{A Context-theoretic Framework for Distributional Semantics}

\author{Daoud Clarke\thanks{Metrica, 140 Old Street, London. E-mail: daoud@metrica.net.}}
\affil{University of Hertfordshire}
\runningauthor{Daoud Clarke}

\usepackage{graphs}
\usepackage{pstricks}
 \usepackage[leqno]{amsmath}

\newcommand{\R}{\mathbb{R}}
 \newcommand{\cont}[1]{\widehat{\mathit{#1}}}

\begin{document}

\maketitle

\section{Introduction}

In recent years, the abundance of text corpora and computing power has allowed
the development of techniques to analyse statistical properties of words. For example
techniques such as latent semantic analysis \cite{Deerwester:90} and its variants,
and measures of distributional similarity \cite{Lin:98,Lee:99} attempt to derive aspects
of the meanings of words by statistical analysis, while statistical information is often
used when parsing to determine sentence structure \cite{Collins:97}. These techniques
have proved useful in many applications within computational linguistics and natural
language processing \cite{Schutze:98,McCarthy:04,Grefenstette:94,Lin:03,Bellegarda:00,Choi:01},
arguably providing evidence that they capture something about
the nature of words that should be included in representations of their meaning.
However, it is very difficult to reconcile these techniques
with existing theories of meaning
in language, which revolve around logical and ontological representations. The new
techniques, almost without exception, can be viewed as dealing with vector-based
representations of meaning, placing meaning (at least at the word level) within the
realm of mathematics and algebra; conversely the older theories of meaning dwell in
the realm of logic and ontology. It seems there is no unifying theory of meaning to
provide guidance to those making use of the new techniques.

The problem appears to be a fundamental one in computational linguistics since
the whole foundation of meaning seems to be in question. The older, logical theories
often subscribe to a model-theoretic philosophy of meaning \cite{Kamp:93,Blackburn:05}
According to this approach, sentences should be translated to
a logical form that can be interpreted as a description of the state of the world. The new
vector-based techniques, on the other hand, are often closer in spirit to the philosophy
of ``meaning as context'', that the meaning of an expression is determined by how it is
used. This is an old idea with origins in the philosophy of \namecite{Wittgenstein:53}, who said
that ``meaning just is use'' and \namecite{Firth:57}, ``You shall know a word by the company
it keeps'', and the distributional hypothesis of \namecite{Harris:68}, that words will occur in
similar contexts if and only if they have similar meanings. This hypothesis is justified by the success of techniques such as latent semantic analysis as well as experimental evidence \cite{Miller:91}. Whilst the two philosophies
are not obviously incompatible --- especially since the former applies mainly at the
sentence level and the latter mainly at the word level --- it is not clear how they relate to
each other.


The problem of how to compose vector representations of meanings of words has recently received increased attention \cite{Widdows:08,Clark:08,Mitchell:08,Erk:09,Preller:09,Guevara:11,Baroni:10} although the problem has been considered in earlier work \cite{Smolensky:90,Landauer:97,Foltz:98,Kintsch:01}. A solution to this problem would have practical as well as philosophical benefits. Current techniques such as latent semantic analysis work well at the word level, but we cannot extend them much beyond this, to the phrase or sentence level, without quickly encountering the data-sparseness problem: there are not enough occurrences of strings of words to determine what their vectors should be merely by looking in corpora. If we knew how such vectors should compose then we would be able to extend the benefits of the vector based techniques to the many applications that require reasoning about the meaning of phrases and sentences.

This paper describes the results of our own efforts to identify a theory that can unite
these two paradigms, and includes a summary of work described in the author's DPhil thesis \cite{Clarke:07}. In addition, we also discuss the relationship between this theory and methods of composition that have recently been proposed in the literature, showing that many of them can be considered as falling within our framework.

Our approach in identifying the framework is summarised in Figure \ref{approach}:
\begin{figure}
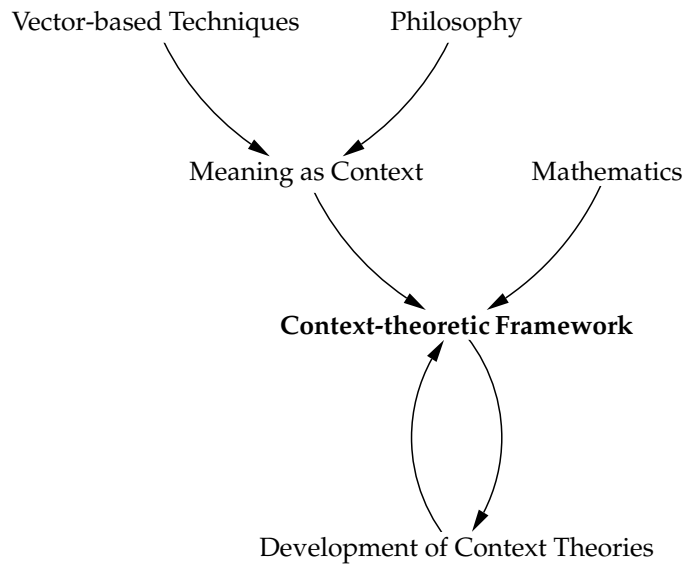

\begin{center}
\begin{graph}(8,8)(0,.5)
\newcommand{\ntext}[4]{\textnode{#1}(#2,#3){#4}[\graphlinecolour{1}]}
\ntext{Mean}{2}{6}{Meaning as Context}
\ntext{Exist}{0}{8}{Vector-based Techniques}
\ntext{Phil}{4}{8}{Philosophy}
\ntext{Math}{6}{6.05}{Mathematics}
\ntext{Cont}{4}{4}{\textbf{Context-theoretic Framework}}
\ntext{Dev}{4}{1}{Development of Context Theories}
\dirbow{Phil}{Mean}{0.1}
\dirbow{Exist}{Mean}{-0.1}
\dirbow{Math}{Cont}{0.1}
\dirbow{Mean}{Cont}{-0.1}
\dirbow{Cont}{Dev}{0.2}
\dirbow{Dev}{Cont}{0.2}
\end{graph}
\end{center}
\caption{Method of Approach in developing the Context-theoretic Framework.}
\label{approach}
\end{figure}
\begin{itemize}
\item Inspired by the philosophy of meaning as context and vector based
         techniques we developed a mathematical model of meaning as context, in
         which the meaning of a string is a vector representing contexts in which
         that string occurs in a hypothetical infinite corpus.
\item The theory on its own is not useful when applied to real world corpora
         because of the problem of data sparseness. Instead we examine the
         mathematical propertes of the model, and abstract them to form a
         framework which contains many of the properties of the model.
         Implementations of the framework are called context theories since they can
         be viewed as theories about the contexts in which strings occur. By
         analogy with the term ``model-theoretic'' we use the term
         ``context-theoretic'' for concepts relating to context theories, thus we call
         our framework the context-theoretic framework.
\item In order to ensure that the framework was practically useful, context
         theories were developed in parallel with the framework itself. The aim
         was to be able to describe existing approaches to representing meaning
         within the framework as fully as possible.
\end{itemize}
In developing the framework we were looking for specific properties; namely, we
wanted it to:
\begin{itemize}
\item provide some guidelines describing in what way the representation of a
         phrase or sentence should relate to the representations of the individual
         words as vectors;
\item require information about the probability of a string of words to be
         incorporated into the representation;
\item provide a way to measure the degree of entailment between strings based
         on the particular meaning representation;
\item be general enough to encompass logical representations of meaning;
\item be able to incorporate the representation of ambiguity and uncertainty,
         including statistical information such as the probability of a parse or the
         probability that a word takes a particular sense.
\end{itemize}
The framework we present is abstract, and hence does not subscribe to a particular method for obtaining word vectors: they may be raw frequency counts, or vectors obtained by a method such as latent semantic analysis.
     Nor does the framework provide a recipe for how to represent meaning in natural language, instead it provides restrictions on the set of possibilities. The advantage
of the framework is in ensuring that techniques are used in a way that is well-founded
in a theory of meaning. For example, given vector representations of words, there is
not one single way of combining these to give vector representations of phrases and
sentences, but in order to fit within the framework there are certain properties of the
representation that need to hold. Any method of combining these vectors in which these
properties hold can be considered within the framework and is thus justified according
to the underlying theory; in addition the framework instructs us as to how to measure
the degree of entailment between strings according to that particular method. We will
attempt to show the broad applicability of the framework by applying it to problems in
natural language processing.

The contribution of this paper is as follows:
\begin{itemize}
\item We define the context-theoretic framework and introduce the mathematics necessary to understand it. The description presented hear is cleaner than that of \cite{Clarke:07}, and in addition we provide examples which should provide intuition for the concepts we describe.
\item We relate the framework to methods of composition that have been proposed in the literature, namely:
\begin{itemize}
\item vector addition \cite{Landauer:97,Foltz:98}
\item the tensor product \cite{Smolensky:90,Clark:07,Widdows:08}
\item the multiplicative models of \namecite{Mitchell:08}
\item matrix multiplication \cite{Rudolph:10,Baroni:10}
\item the approach of \namecite{Clark:08}.
\end{itemize}
\end{itemize}

\section{Context Theory}

In this section, we define the fundamental concept of our concern, a \textbf{context theory} and discuss its properties.

\begin{definition}[Context Theory]
A context theory is a tuple $\langle A, \mathcal{A}, \xi, V, \psi \rangle$, where $A$ is a set (the alphabet), $\mathcal{A}$ is a unital algebra over the real numbers, $\xi$ is a function from $A$ to $\mathcal{A}$, $V$ is an abstract Lebesgue space and $\psi$ is an injective linear map from $\mathcal{A}$ to $V$.
\end{definition}

We will explain each part of this definition, introducing the necessary mathematics as we proceed. We assume the reader is familiar with linear algebra; see \cite{Halmos:74} for definitions that are not included here.

\subsection{Algebra over a field}

We have identified an algebra over a field as an important construction since it generalises nearly all the methods of vector-based composition that have been proposed.

\begin{definition}[Algebra over a field]
An algebra over a field (or simply \textbf{algebra} when there is no ambiguity) is a vector space $\mathcal{A}$ over a field $K$ together with a binary operation $(a,b)\mapsto ab$ on $\mathcal{A}$ that is bilinear, i.e.
\begin{align*}
a(\alpha b + \beta c) &= \alpha ab + \beta ac\\
(\alpha a+\beta b)c &= \alpha ac + \beta bc
\end{align*}
and associative, i.e. $(ab)c = a(bc)$ for all $a,b,c\in \mathcal{A}$ and all $\alpha,\beta \in K$.\footnote{Some authors do not place the requirement that an algebra is associative, in which case our definition would refer to an \emph{associative algebra}.} An algebra is called \textbf{unital} if it has a distinguished \textbf{unity} element $1$ satisfying $1x = x1 = x$ for all $x\in\mathcal{A}$.
\end{definition}
We are generally only interested in \textbf{real} algebras, i.e.~the situation where $K$ is the field of real numbers, $\mathbb{R}$.

\begin{example}
The square real-valued matrices of order $n$ form a real unital associative algebra under standard matrix multiplication. The vector operations are defined entry-wise. The unity element of the algebra is the identity matrix.
\end{example}

This means that our proposal is more general than that of \namecite{Rudolph:10}, who suggest using matrix multiplication as a framework for distributional semantic composition. The main differences in our proposal are:
\begin{itemize}
\item We allow dimensionality to be infinite, instead of restricting ourselves to finite-dimensional matrices;
\item Matrix algebras form a $*$-algebra, whereas we do not currently place this requirement;
\item We emphasise the order structure that is inherent in real vector spaces when there is a distinguished basis.
\end{itemize}

The purpose of $\xi$ in the context theory is to associate elements of the algebra with strings of words. Considering only the multiplication of $\mathcal{A}$ (and ignoring the vector operations), $\mathcal{A}$ is a monoid, since we assumed that the multiplication on $\mathcal{A}$ is associative. Then $\xi$ induces a monoid homomorphism $a\mapsto \hat{a}$ from $A^*$ to $\mathcal{A}$. We denote the mapped value of $a\in A^*$ by $\hat{a}\in\mathcal{A}$, which is defined as follows:
$$\hat{a} = \xi(a_1)\xi(a_2)\ldots\xi(a_n)$$
where $a = a_1a_2\ldots a_n$ for $a_i\in A$, and we define $\hat{\epsilon} = 1$, where $\epsilon$ is the empty string. Thus, the mapping defined by $\hat{\ }$ allows us to associate an element of the algebra with every string of words.

The algebra is what tells us how meanings compose. A crucial part of our thesis is that meanings can be represented by elements of an algebra, and that the type of composition that can be defined using an algebra is general enough to describe the composition of meaning in natural language. To go some way towards justifying this, we give several examples of algebras that describe methods of composition that have been proposed in the literature: namely point-wise multiplication \cite{Mitchell:08}, vector addition \cite{Landauer:97,Foltz:98} and the tensor product \cite{Smolensky:90,Clark:07,Widdows:08}.

\begin{table}
\begin{center}
\begin{tabular}{llll}
\hline
 & $d_1$ & $d_2$ & $d_3$\\
\hline
 \textit{cat} & 0 & 2 & 3\\
 \textit{animal} & 2 & 1 & 2\\
 \textit{big} & 1 & 3 & 0 \\
\hline
\end{tabular}
\end{center}
\label{catdog}
\caption{Example of possible occurrences for three terms in three different contexts.}
\end{table}

\begin{example}[Point-wise multiplication]
Consider the $n$-dimensional real vector space $\R^n$. We describe a vector $u \in \R^n$ in terms of its components as $(u_1,u_2,\ldots u_n)$ with each $u_i\in\R$. We can define a multiplication $\cdot$ on this space by
$$(u_1,u_2,\ldots, u_n)\cdot(v_1,v_2,\ldots, v_n) = (u_1v_1, u_2v_2, \ldots u_nv_n)$$
It is easy to see that this satisfies the requirements for an algebra specified above. Table \ref{catdog} shows a simple example of possible occurrences for three terms in three different contexts, $d_1$,$d_2$ and $d_3$ which may, for example, represent documents.
We use this to define the mapping $\xi$ from terms to vectors. Thus, in this example, we have $\xi(cat) = (0,2,3)$ and $\xi(big) = (1,3,0)$. Under point-wise multiplication, we would have
$$\cont{big\ cat} = \xi(big)\cdot\xi(cat) = (1,3,0)\cdot(0,2,3) = (0,6,0).$$
\end{example}

One commonly used operation for composing vector-based representations of meaning is vector addition. As noted by \namecite{Rudolph:10}, this can be described using matrix multiplication, by embedding an $n$-dimensional vector $u$ into a matrix of order $n+1$:
$$\left(\begin{array}{lllll}
\alpha\rule{0.15cm}{0cm}  & u_1 & u_2 & \cdots & u_n \\
0 & \alpha & 0 &\cdots & 0 \\
0 & 0 & \alpha & \cdots & 0 \\
\vdots & \vdots & \vdots & &\vdots \\
0 & 0 & 0 &\cdots & \alpha\\
\end{array} \right)$$
where $\alpha = 1$. The set of all such matrices, for all real values of $\alpha$, forms a \textbf{subalgebra} of the algebra of matrices of order $n+1$. A subalgebra of an algebra $\mathcal{A}$ is a sub-vector space of $\mathcal{A}$ which is closed under the multiplication of $\mathcal{A}$. This subalgebra can be equivalently described as follows:
\begin{example}[Additive algebra]
For two vectors $u = (\alpha,u_1,u_2,\ldots u_n)$ and $v = (\beta,v_1,v_2\ldots v_n)$ in $\R^{n+1}$, we define the additive product $\boxplus$ by
$$u\boxplus v = (\alpha \beta,\ \alpha v_1 + \beta u_1,\ \alpha v_2 + \beta u_2,\ \ldots \alpha v_n + \beta u_n)$$
To verify that this multiplication makes $\R^{n+1}$ an algebra, we can directly verify the bilinear and associativity requirements, or check that it is isomorphic to the subalgebra of matrices discussed above.

Using the table from the previous example, we define $\xi_+$ so that it maps $n$-dimensional context vectors to $\R^{n+1}$, where the first component is $1$, so $\xi_+(\textit{big}) = (1,1,3,0)$ and $\xi_+(\textit{cat}) = (1,0,2,3)$ and
$$\cont{big\ cat} = \xi_+(\textit{big})\boxplus \xi_+(\textit{cat}) = (1,1,5,3).$$
\end{example}

Point-wise multiplication and addition are not attractive as methods for composing meaning in natural language since they are commutative, whereas natural language is inherently non-commutative. One obvious method of composing vectors that is not commutative is the \textbf{tensor product}. This method of composition can be viewed as a product in an algebra by considering the \textbf{tensor algebra}, which is formed from direct sums of all tensor powers of a base vector space.

We assume the reader is familiar with the tensor product and direct sum (see \cite{Halmos:74} for definitions); we recall their basic properties here. Let $V_n$ denote a vector space of dimensionality $n$ (note that all vector spaces of a fixed dimensionality are isomorphic). Then the tensor product space $V_n \otimes V_m$ is isomorphic to a space $V_{nm}$ of dimensionality $nm$; moreover given orthonormal bases $B = \{b_1, b_2,\ldots, b_n\}$ for $V_n$ and $C = \{c_1, c_2,\ldots, c_m\}$ for $V_m$ there is an orthonormal basis for $V_{nm}$ defined by
$$\{b_i\otimes c_j : 1\le i \le n \text{ and } 1 \le j \le m\}.$$

\begin{example}
The multiplicative models of \cite{Mitchell:08} correspond to the class of finite dimensional algebras. Let $\mathcal{A}$ be a finite-dimensional vector space. Then every associative bilinear product on $\mathcal{A}$ can be described by a linear function $T$ from $\mathcal{A}\otimes \mathcal{A}$ to $\mathcal{A}$, as required in Mitchell and Lapata's model. To see this, consider the action of the product $\cdot$ on two orthonormal basis vectors $a$ and $b$ of $\mathcal{A}$. This is a vector in $\mathcal{A}$, thus we can define $T(a\otimes b) = a \cdot b$. By considering all basis vectors, we can define the linear function $T$.
\end{example}

If the tensor product can loosely be viewed as ``multiplying'' vector spaces, then the direct sum is like adding them; the space $V_n \oplus V_m$ has dimensionality $n + m$ and has basis vectors
$$\{b_i\oplus 0 : 1\le i \le n\} \cup \{0 \oplus c_j : 1 \le j \le m\};$$
it is usual to write $b \oplus 0$ as $b$ and $0 \oplus c$ as $c$.

\begin{example}[Tensor algebra]
If $V$ is a vector space, then we define $T(V)$, the \textbf{free algebra} of \textbf{tensor algebra} generated by $V$ as:
$$T(V) = \mathbb{R}\oplus V \oplus (V\otimes V) \oplus (V\otimes V\otimes V) \oplus \cdots$$
where we assume that the direct sum is commutative. We can think of it as the direct sum of all tensor powers of $V$, with $\R$ representing the zeroth power. In order to make this space an algebra, we define the product on elements of these tensor powers, viewed as subspaces of the tensor algebra, as their tensor product. This is enough to define the product on the whole space, since every element can be written as a sum of tensor powers of elements of $V$. There is a natural embedding from $V$ to $T(V)$, where each element maps to an element in the first tensor power. Thus for example we can think of $u$, $u\otimes v$, and $u\otimes v + w$ as elements of $T(V)$, for all $u,v,w\in V$. 

This product defines an algebra since the tensor product is a bilinear operation. Taking $V = \R^3$ and using $\xi$ as the natural embedding from the context vector of a string $T(V)$, our previous example becomes
\begin{eqnarray*}
\cont{big\ cat} &=& \xi(big)\otimes\xi(cat)\\
 &=& (1,3,0)\otimes(0,2,3)\\
  &\cong & (1(0,2,3),3(0,2,3),0(0,2,3))\\
   &\cong & (0,2,3,0,6,9,0,0,0)
\end{eqnarray*}
where the last two lines demonstrate how a vector in $\R^3\otimes \R^3$ can be described in the isomorphic space $\R^9$.
\end{example}


\subsection{Vector lattices}

The next part of the definition specifies an abstract Lebesgue space. This is a special kind of vector lattice,  or even more generally, a partially ordered vector space.

\begin{definition}[Partially ordered vector space]\index{vector space!partially ordered|textbf}
A partially ordered vector space $V$ is a real vector space together with a partial ordering $\le$ such that:
\vspace{0.1cm}\\
\indent if $x \le y$ then $x + z \le y + z$\\
\indent if $x \le y$ then $\alpha x \le \alpha y$
\vspace{0.1cm}\\
for all $x,y,z \in V$, and for all $\alpha \ge 0$. Such a partial ordering is called a \textbf{vector space order} on $V$. An element $u$ of $V$ satisfying $u \ge 0$ is called a \textbf{positive element}; the set of all positive elements of $V$ is denoted $V^+$. If $\le$ defines a lattice on $V$ then the space is called a \textbf{vector lattice} or \textbf{Riesz space}.
\end{definition}

\begin{figure}
\begin{center}
\input{orangefruit2.pst}
\caption{Vector representations of the terms \emph{orange} and \emph{fruit} based on hypothetical occurrences in six documents and their vector lattice meet (the darker shaded area).}
\label{orangefruit}
\end{center}
\end{figure}

\begin{example}[Lattice operations on $\R^n$]
A vector lattice captures many properties that are inherent in real vector spaces when there is a \emph{distinguished basis}. In $\R^n$, given a specific basis, we can write two vectors $u$ and $v$ as sequences of numbers: $u = (u_1,u_2,\ldots u_n)$ and $v = (v_1,v_2,\ldots v_n)$. This allows us to define the lattice operations of meet $\land$ and join $\lor$ as
\begin{eqnarray*}
u\land v &=& (\min(u_1,v_1),\min(u_2,v_2),\ldots \min(u_n,v_n))\\
u\lor v &=& (\max(u_1,v_1),\max(u_2,v_2),\ldots \max(u_n,v_n))
\end{eqnarray*}
i.e.~the component-wise minimum and maximum, respectively. A graphical depiction of the meet operation is shown in figure \ref{orangefruit}.
\end{example}
The vector operations of addition and multiplication by scalar, which can be defined in a similar component-wise fashion, are nevertheless independent of the particular basis chosen. This makes them particularly suited to physical applications, where it is often a requirement that there is no preferred direction. Conversely, the lattice operations depend on the choice of basis, so the operations as defined above would behave differently if the components were written using a different basis. We argue that it makes sense for us to consider these properties of vectors in the context of computational linguistics since we can often have a distinguished basis: namely the one defined by the contexts in which terms occur. Of course it is true that techniques such as latent semantic analysis introduce a new basis which does not have a clear interpretation in relation to contexts; nevertheless they nearly always identify a distinguished basis which we can use to define the lattice operations.

We argue that the mere association of words with vectors is not enough to constitute a theory of meaning. Vector representations allow the measurement of similarity or distance, through an inner product or metric, however we believe it is also important for a theory of meaning to model \emph{entailment}, a relation which plays an important r\^ole in logical theories of meaning. In propositional and first order logic, the entailment relation is a partial ordering, in fact it is a Boolean algebra, which is a special kind of lattice. It seems natural to consider whether the lattice structure that is inherent in the vector representations used in computational linguistics can be used to model entailment.

We believe our framework is suited to all vector-based representations of natural language meaning, however the vectors are obtained. Given this assumption, we can only justify our assumption that the partial order structure of the vector space is suitable to represent the entailment relation by observing that it has the right kind of properties we would expect from this relation.

There may, however, be more justification for this assumption, based on the case where the vectors for terms are simply their frequencies of occurrences in $n$ different contexts, so that they are vectors in $\R^n$. In this case, the relation $\xi(x) \le \xi(y)$ means that $y$ occurs at least as frequently as $x$ in every context. This means that $y$ occurs in at least as wide a range of contexts as $x$, and occurs as least as frequently as $x$. Thus the statement ``$x$ entails $y$ if and only if $\xi(x) \le \xi(y)$'' can be viewed as a stronger form of the distributional hypothesis of \namecite{Harris:68}.

In fact, this idea can be related to the notion of ``distributional generality'', introduced by \namecite{Weeds:04} (see also \cite{Geffet:05}). A term $x$ is distributionally more general than another term $y$ if $x$ occurs in a subset of the contexts that $y$ occurs in. The idea is that distributional generality may be connected to \emph{semantic generality}. An example of this is the hypernymy or ``is a'' relation that is used to express generality of concepts in ontologies, for example, the term \emph{animal} is a hypernym of \emph{dog} since a dog is an animal.
They explain the connection to distributional generality as follows:
\begin{quote}
Although one can obviously think of counter-examples, we would generally expect that the more specific term \emph{dog} can only be used in contexts where \emph{animal} can be used and that the more general term \emph{animal} might be used in all of the contexts where \emph{dog} is used and possibly others. Thus, we might expect that distributional generality is correlated with semantic generality\ldots
\end{quote}

Our proposal, in the case where words are represented by frequency vectors, can be considered a stronger version of distributional generality, where the additional requirement is on the frequency of occurrences. In practice, this assumption is unlikely to be compatible with the ontological view of entailment. For example the term \emph{entity} is semantically more general than the term \emph{animal}, however \emph{entity} is unlikely to occur more frequently in each context, since it is a rarer word. A more realistic foundation for this assumption might be if we were to consider the components for a word to represent the plausibility of observing the word in each context. The question then of course, is how such vectors might be obtained. Another possibility is to attempt to weight components in such a way that entailment becomes a plausible interpretation for the partial ordering relation.

Even if we allow for such alternatives, however, in general it is unlikely that the relation will hold between any two strings, since $u \le v$ if and only if $u_i \le v_i$ for each component, $u_i,v_i$, of the two vectors. Instead, we propose to allow for \emph{degrees of entailment}. We take a Bayesian perspective on this, and suggest that the degree of entailment should take the form of a conditional probability. In order to define this, however, we need some additional structure on the vector lattice that allows it to be viewed as a description of probability, by requiring it to be an ``abstract Lebesgue space''.


\begin{definition}[Banach lattice]
A Banach lattice $V$ is a vector lattice together with a norm $\|\cdot\|$ such that $V$ is complete with respect to $\|\cdot\|$.  
\end{definition}

\begin{definition}[Abstract Lebesgue Space]
An Abstract Lebesgue (or AL) space is a Banach lattice $V$ such that 
$$\|u + v\| = \|u\| + \|v\|$$
for all $u,v$ in $V$ with $u \ge 0$, $v \ge 0$ and $u\land v = 0$.
\end{definition}

\begin{example}[$\ell^p$ spaces]
Let $u = (u_1, u_2, \ldots)$ be an infinite sequence of real numbers. We can view $u_i$ as components of the infinite-dimensional vector $u$. We call the set of all such vectors the \textbf{sequence space}; it is a vector space where the operations are defined component-wise. We define a set of norms, the $\ell^p$-norms, on the space of all such vectors by
$$\|u\|_p = \left( \sum_{i>0} |u_i|^p \right)^{1/p}$$
The space of all vectors $u$ for which $\|u\|_p$ is finite is called the $\ell^p$ space. Considered as vector spaces, these are Banach spaces, since they are complete with respect to the associated norm, and under the component-wise lattice operations, they are Banach lattices. In particular, the $\ell^1$ space is an abstract Lebesgue space under the $\ell^1$ norm.
\end{example}

The finite-dimensional real vector spaces $\R^n$ can be considered as special cases of the sequence spaces (consisting of vectors in which all but $n$ components are zero) and, since they are finite-dimensional, we can use any of the $\ell^p$ norms. Thus, our previous examples, in which $\xi$ mapped terms to vectors in $\R^n$ can be considered as mapping to abstract Lebesgue spaces, if we adopt the $\ell^1$ norm.

\subsection{Degrees of entailment}

An abstract Lebesgue space has many of the properties of a measure space, where the set operations of a measure space are replaced by the lattice operations of the vector space. This means that we can think of an abstract Lebesgue space as a vector-based probability space. Here, events correspond to positive elements with norm less than or equal to 1; the probability of an event $u$ is given by the norm (which we shall always assume is the $\ell^1$ norm), and the joint probability of two events $u$ and $v$ is $\|u\land v\|_1$.

\begin{definition}[Degree of entailment]
We consider the degree to which $u$ entails $v$ to be the conditional probability of $v$ given $u$:
$$\mathrm{Ent}(u,v) = \frac{\| u \land v\|_1}{\|u\|_1}.$$
\end{definition}
If we are only interested in degrees of entailment (i.e.~conditional probabilities) and not probabilities, then we can drop the requirement that the norm should be less than or equal to one, since conditional probabilities are automatically normalised. This definition, together with the multiplication of the algebra, allows us to compute the degree of entailment between any two strings according to the context theory.

\begin{example}
The vectors given in Table \ref{catdog} give the following calculation for the degree to which \emph{cat} entails \emph{animal}:
\begin{eqnarray*}
\xi(cat) &=& (0,2,3)\\
\xi(animal) &=& (2,1,2)\\
\xi(cat)\land\xi(animal) &=& (0,1,2)\\
\mathrm{Ent}(\xi(cat),\xi(animal)) &=& \|\xi(cat)\land\xi(animal)\|_1/\|\xi(cat)\|_1 = 3/5
\end{eqnarray*}
\end{example}

An important question is how this context-theoretic definition of the degree of entailment relates to more familiar notions of entailment.\footnote{Thanks are due to the anonymous reviewer who identified this question and related issues.} There are three main ways in which the term entailment is used:
\begin{itemize}
\item The \textbf{model-theoretic} sense of entailment in which a theory $A$ entails a theory $B$ if every model of $A$ is also a model of $B$. It was shown in \cite{Clarke:07} that this type of entailment can be described using context theories, where sentences are represented as projections on a vector space.
\item Entailment between terms in the word net hierarchy, for example the \textbf{hypernymy} or \textbf{is-a} relation between the terms \emph{cat} and \emph{animal} encodes the fact that a cat is an animal. In \cite{Clarke:07} we showed that such relations can be encoded in the partial order structure of a vector lattice.
\item Human common-sense judgments as to whether one sentence entails or implies another sentence, as used in the Recognising Textual Entailment Challenges \cite{Dagan:05}.
\end{itemize}
Our context-theoretic notion of entailment is thus intended to generalise both the first two senses of entailment above. In addition, we hope that context theories will be useful in the practical application of recognising textual entailment.

Our definition is more general than the model-theoretic and hypernymy notions of entailment however, as it allows the measurement of a degree of entailment between any two strings: as an extreme example, one may measure the degree to which ``not a'' entails ``in the''. Whilst this may not be useful or philosophically meaningful, we view it as a practical consequence of the fact that every string has a vector representation in our model, which coincides with the current practice in vector-based compositionality techniques.

\subsection{Lattice ordered algebras}


A large class of context theories make use of a \textbf{lattice ordered algebra} which merges the lattice ordering of the vector space $V$ with the product of $\mathcal{A}$.
\begin{definition}[Partially ordered algebra]
A partially ordered algebra $\mathcal{A}$ is an algebra which is also a partially ordered vector space, which satisfies $u\cdot v \ge 0$ for all $u,v\in\mathcal{A}^+$. If the partial ordering is a lattice, then $\mathcal{A}$ is called a \textbf{lattice ordered algebra}.
\end{definition}
\begin{example}[Lattice ordered algebra of matrices]
The matrices of order $n$ form a lattice ordered algebra under normal matrix multiplication, where the lattice operations are defined as the entry-wise minimum and maximum.
\end{example}
\begin{example}[Operators on $\ell^p$ spaces]
\label{riesz}
Operators on the $\ell^p$ spaces are also lattice ordered algebras, by the Riesz-Kantorovich theorem \cite{Abramovich:02}, with the operations defined by:
\begin{eqnarray*}
(S \lor T)(u) & = & \sup \{S(v) + T(w): v,w\in U^+ \text{ and } v + w = u\}\\
(S \land T)(u) & = & \inf \{S(v) + T(w): v,w\in U^+ \text{ and } v + w = u\}
\end{eqnarray*}
\end{example}

%

If $\mathcal{A}$ is a lattice ordered algebra which is also an abstract Lebesgue space, then $\langle A, \mathcal{A}, \xi, \mathcal{A}, 1 \rangle$ is a context theory. Many of the examples we discuss will be of this form, so we will use the shorthand notation, $\langle A, \mathcal{A}, \xi\rangle$. It is tempting to adopt this as the definition of context theory, however, as we will see, this is not supported by our prototypical example of a context theory (which we will introduce in the next section) as in this case the algebra is not necessarily lattice ordered.

\section{Context Algebras}

In this section we describe the prototypical examples of a context theory, the \textbf{context algebras}. The definition of a context algebra originates in the idea that the notion of ``meaning as context'' can be extended beyond the word level to strings of arbitrary length. In fact, the notion of context algebra can be thought of as a generalisation of the \textbf{syntactic monoid} of a formal language: instead of a set of strings defining the language, we have a \emph{fuzzy set} of strings, or more generally, a real-valued function on a free monoid.

\begin{definition}[Real-valued language]
Let $A$ be a finite set of symbols. A \textbf{real-valued language} (or simply a \textbf{language} when there is no ambiguity) $L$ on $A$ is a function from $A^*$ to $\R$. If the range of $L$ is a subset of $\R^+$ then $L$ is called a \textbf{positive language}. If the range of $L$ is a subset of $[0,1]$ then $L$ is called a \textbf{fuzzy language}. If $L$ is a positive language such that $\sum_{x\in A^*} L(x) = 1$ then $L$ is a \textbf{probability distribution} over $A^*$.
\end{definition}
The following inclusion relation applies amongst these classes of language:
$$\text{ distribution }\Longrightarrow \text{ fuzzy }\Longrightarrow \text{ positive }\Longrightarrow \text{ real-valued }$$

Since $A^*$ is a countable set, the set $\R^{A^*}$ of functions from $A^*$ to $\R$ is isomorphic to the sequence space, and we shall treat them equivalently. We denote by $\ell^p(A^*)$ the set of functions with a finite $\ell^p$ norm, when considered as sequences. There is another heirarchy of spaces given by the inclusion of the $\ell^p$ spaces: $\ell^p(A^*) \subseteq \ell^q(A^*)$ if $p \le q$. In particular,
$$\ell^1(A^*) \subseteq \ell^2(A^*) \subseteq \ell^\infty(A^*)\subseteq \R^{A^*}$$
where the $\ell^\infty$ norm gives the maximum value of the function and $\ell^\infty(A^*)$ is the space of all bounded functions on $A^*$.

Note that probability distributions are in $\ell^1(A^*)$ and fuzzy languages are in $\ell^\infty(A^*)$. If $L \in \ell^1(A^*)^+$ (the space of positive functions on $A^*$ such that the sum of all values of the function is finite) then we can define a probability distribution $p_L$ over $A^*$ by $p_L(x) = L(x)/\|L\|_1$. Similarly, if $L \in \ell^\infty(A^*)^+$ (the space of bounded positive functions on $A^*$) then we can define a fuzzy language $f_L$ by $f_L(x) = L(x)/\|L\|_\infty$.

\begin{example}
Given a finite set of strings $C \subset A^*$, which we may imagine to be a corpus of documents, define $L(x) = 1/|C|$ if $x \in C$, or $0$ otherwise. Then $L$ is a probability distribution over $A^*$.
\end{example}

\begin{example}
Let $L$ be a language such that $L(x) = 0$ for all but a finite subset of $A^*$. Then $L \in \ell^p(A^*)$ for all $p$.
\end{example}

\begin{example}
Let $L$ be the language defined by $L(x) = |x|$ where $x$ is the length of (i.e.~number of symbols in) string $x$. Then $L$ is a positive language which is not bounded: for any string $y$ there exists a $z$ such that $L(z) > L(y)$, for example $z = ay$ for $a \in A$.
\end{example}
\begin{example}
Let $L$ be the language defined by $L(x) = 1/2$ for all $x$. Then $L$ is a fuzzy language but $L\notin \ell^1(A^*)$
\end{example}

We will assume now that $L$ is fixed, and consider the properties of contexts of strings with respect to this language.
\begin{definition}[Context vectors]
Let $L$ be a language on $A$. For $x\in A^*$, we define the context of $x$ as a vector $\hat{x}\in \R^{A^*\times A^*}$, i.e.~a real-valued function on pairs of strings:
$$\hat{x}(y,z) = L(yxz).$$
\end{definition}
Our thesis is centred around these vectors, and it is their properties that form the inspiration for the context-theoretic framework.

The question we are addressing is: does there exist some algebra $\mathcal{A}$ containing the context vectors of strings in $A^*$ such that $\hat{x}\cdot \hat{y} = \widehat{xy}$ where $x,y\in A^*$ and $\cdot$ indicates multiplication in the algebra? As a first try, consider the vector space $L^\infty(A^*\times A^*)$ in which the context vectors live. Is it possible to define multiplication on the whole vector space such that the condition just specified holds?
\begin{example}
Consider the language $C$ on the alphabet $A = \{a,b,c,d,e,f\}$ defined by $C(abcd) = C(aecd) = C(abfd) = \frac{1}{3}$ and $C(x) = 0$ for all other $x \in A^*$. Now if we take the shorthand notation of writing the basis vector in $L^\infty(A^*\times A^*)$ corresponding to a pair of strings as the pair of strings itself then
\begin{eqnarray*}
\hat{b} &=& \tfrac{1}{3}(a,cd) +  \tfrac{1}{3}(a,fd)\\
\hat{c} &=& \tfrac{1}{3}(ab,d) +  \tfrac{1}{3}(ae,d)\\
\widehat{bc} &=& \tfrac{1}{3}(a,d)
\end{eqnarray*}
%
%
It would thus seem sensible to define multiplication of contexts so that $ \tfrac{1}{3}(a,cd)\cdot  \tfrac{1}{3}(ab,d) =  \tfrac{1}{3}(a,d)$. However we then find
$$\hat{e}\cdot \hat{f} =  \tfrac{1}{3}(a,cd)\cdot  \tfrac{1}{3}(ab,d) \neq \widehat{ef} = 0$$
showing that this definition of multiplication doesn't provide us with what we are looking for. In fact, if there did exist a way to define multiplication on contexts in a satisfactory manner it would necessarily be far from intuitive, as, in this example, we would have to define $(a,cd)\cdot (ab,d) = 0$ meaning the product $\hat{b}\cdot\hat{c}$ would have to have a non-zero component derived from the products of context vectors $(a,fd)$ and $(ae,d)$ which don't relate at all to the contexts of $bc$. This leads us to instead define multiplication on a subspace of  $L^\infty(A^*\times A^*)$.
\end{example}
\begin{definition}[Generated Subspace $\mathcal{A}$]
The subspace $\mathcal{A}$ of $L^\infty(A^*\times A^*)$ is the set defined by
$$\mathcal{A} = \{a : a = \sum_{x\in A^*}\alpha_x \hat{x}\text{ for some }\alpha_x \in \R\}$$
\end{definition}

Because of the way we define the subspace, there will always exist some basis\index{basis} $\mathcal{B} = \{\hat{u} : u \in B\}$ where $B \subseteq A^*$, and we can define multiplication on this basis by $\hat{u}\cdot\hat{v} = \widehat{uv}$ where $u,v \in B$. Defining multiplication on the basis defines it for the whole vector subspace, since we define multiplication to be linear, making $\mathcal{A}$ an algebra.

However there are potentially many different bases we could choose, each corresponding to a different subset of $A^*$, and each giving rise to a different definition of multiplication. Remarkably, this isn't a problem:

\begin{proposition*}[Context Algebra]
Multiplication on $\mathcal{A}$ is the same irrespective of the choice of basis $B$.
\end{proposition*}
\begin{proof*}
We say $B \subseteq A^*$ defines a basis $\mathcal{B}$ for $\mathcal{A}$ when $\mathcal{B}$ is a basis such that $\mathcal{B} = \{\hat{x}: x\in B\}$. Assume there are two sets $B_1, B_2 \subseteq A^*$ that define corresponding bases $\mathcal{B}_1$ and $\mathcal{B}_2$ for $\mathcal{A}$. We will show that multiplication in basis $\mathcal{B}_1$ is the same as in the basis $\mathcal{B}_2$.

We represent two basis elements $\hat{u}_1$ and $\hat{u}_2$ of $\mathcal{B}_1$ in terms of basis elements of $\mathcal{B}_2$:
$$\hat{u}_1 = \sum_i \alpha_i \hat{v}_i \quad\text{and}\quad
\hat{u}_2 = \sum_j \beta_j \hat{v}_j,$$
for some $u_i \in B_1$, $v_j \in B_2$ and $\alpha_i, \beta_j  \in \R$.
 First consider multiplication in the basis $\mathcal{B}_1$. Note that $\hat{u}_1 = \sum_i \alpha_i \hat{v}_i$ means that $L(xu_1y) = \sum_i \alpha_i L(xv_iy)$ for all $x,y \in A^*$. This includes the special case where $y = u_2y'$ so $$L(xu_1u_2y') = \sum_i \alpha_i L(xv_iu_2y')$$ for all $x, y' \in A^*$.
Similarly, we have $L(xu_2y) = \sum_j \beta_j L(xv_jy)$ for all $x,y \in A^*$ which includes the special case $x = x'v_i$, so $L(x'v_iu_2y) = \sum_j \beta_j L(x'v_iv_jy)$ for all $x',y \in A^*$. Inserting this into the above expression yields
$$L(xu_1u_2y) = \sum_{i,j} \alpha_i\beta_j L(xv_iv_jy)$$
for all $x,y \in A^*$ which we can rewrite as
$$\hat{u}_1\cdot\hat{u}_2 = \widehat{u_1u_2} = \sum_{i,j}\alpha_i\beta_j (\hat{v}_i\cdot\hat{v}_j)
= \sum_{i,j}\alpha_i\beta_j \widehat{v_iv_j}.$$
Conversely, the product of $u_1$ and $u_2$ using the basis $\mathcal{B}_2$ is
$$\hat{u}_1\cdot \hat{u}_2 = \sum_i \alpha_i \hat{v}_i \cdot \sum_j \beta_j \hat{v}_j =  \sum_{i,j}\alpha_i\beta_j (\hat{v}_i\cdot\hat{v}_j)$$
thus showing that multiplication is defined independently of what we choose as the basis.
\end{proof*}

\begin{example}
Returning to the previous example, we can see that in this case multiplication is in fact defined on $L^\infty(A^*\times A^*)$ since we can describe each basis vector in terms of context vectors:
\begin{eqnarray*}
(a,fd)\cdot(ae,d) &=& 3(\hat{b} - \hat{e})\cdot 3(\hat{c} - \hat{f}) = -3(a,d)\\
(a,cd)\cdot(ae,d) &=& 3\hat{e}\cdot 3(\hat{c} - \hat{f}) = 3(a,d)\\
(a,fd)\cdot(ab,d) &=& 3(\hat{b} - \hat{e})\cdot 3\hat{f} = 3(a,d)\\
(a,cd)\cdot(ab,d) &=& 3\hat{e}\cdot 3\hat{f} = 0,
\end{eqnarray*}
thus confirming what we predicted about the product of $\hat{b}$ and $\hat{c}$: the value is only correct because of the negative correction from $(a,fd)\cdot(ae,d)$. This example also serves to demonstrate an important property of context algebras: they do not satisfy the positivity condition; i.e.~it is possible for positive vectors (those with all components greater than or equal to zero) to have a non-positive product. This means they are not necessarily partially ordered algebras under the normal partial order. Compare this to the case of matrix multiplication, for example, where the product of two positive matrices is always positive.
\end{example}

The notion of a context theory is founded on the prototypical example given by context vectors. So far we have shown that multiplication can be defined on the vector space $\mathcal{A}$ generated by context vectors of strings, however we have not discussed the lattice properties of the vector space. In fact, $\mathcal{A}$ does not come with a natural lattice ordering that makes sense for our purposes, however, the original space $\R^{A^*\times A^*}$ does --- it is isomorphic to the sequence space. Thus $\langle A, \mathcal{A}, \xi, \R^{A^*\times A^*},\psi\rangle$ will form our context theory, where $\xi(a) = \hat{a}$ for $a \in A$ and $\psi$ is the \textbf{canonical map} which simply maps elements of $\mathcal{A}$ to themselves, but considered as elements of $\R^{A^*\times A^*}$. There is an important caveat here however: we required that the vector lattice be an abstract Lebesgue space, which means we need to be able to define a norm on it. The $\ell^1$ norm on $\R^{A^*\times A^*}$ is an obvious candidate, however it is not guaranteed to be finite. This is where the nature of the underlying language $L$ becomes important.

We might hope that the most restrictive class of the languages we discussed, the probability distributions over $A^*$ would guarantee that the norm is finite. Unfortunately, this is not the case, as the following example demonstrates.

\begin{example}
Let $L$ be the language defined by
$$L(a^{2^n}) = 1/2^{n+1}$$
for integer $n \ge 0$, and zero otherwise, where by $a^n$ we mean $n$ repetitions of $a$, so for example, $L(a) = \frac{1}{2}$, $L(aa) = \frac{1}{4}$, $L(aaa) = 0$ and $L(aaaa) = \frac{1}{8}$. Then $L$ is a probability distribution over $A^*$, since $L$ is positive and $\|L\|_1 = 1$. However $\|\hat{a}\|_1$ is infinite, since each string $x$ for which $L(x) > 0$ contributes $1/2$ to the value of the norm, and there are an infinite number of such strings.
\end{example}

The problem in the previous example is that the average string length is infinite. If we restrict ourselves to probability distributions over $A^*$ in which the average string length is finite, then the problem goes away.

\begin{proposition*}
Let $L$ be a probability distribution over $A^*$ such that
$$\bar{L} = \sum_{x\in A^*} L(x)|x|$$
is finite, where $|x|$ is the number of symbols in string $x$; we will call such languges \textbf{finite average length} languges. Then $\|\hat{y}\|_1$ is finite for each $y\in A^*$.
\end{proposition*}
\begin{proof*}
Denote the number of occurrences of string $y$ as a substring of string $x$ by $|x|_y$. Clearly $|x|_y \le |x|$ for all $x,y \in A^*$. Moreover,
$$\|\hat{y}\|_1 = \sum_{x\in A^*}L(x)|x|_y \le  \sum_{x\in A^*}L(x)|x|$$
and so $\|\hat{y}\|_1 \le \bar{L}$ is finite for all $y\in A^*$.
\end{proof*}

If $L$ is finite average length, then $\mathcal{A} \subseteq \ell^1(A^*\times A^*)$, and so  $\langle A, \mathcal{A}, \xi, \ell^1(A^*\times A^*),\psi\rangle$ is a context theory, where $\psi$ is the canonical map from $\mathcal{A}$ to $\ell^1(A^*\times A^*)$.
Thus context algebras of finite average length languages provide our prototypical examples of context theories.

\subsection{Discussion}

The benefit of the context-theoretic framework is in providing a space of exploration for models of meaning in language. Our effort has been in finding principles by which to define the boundaries of this space. Each of the key boundaries, namely, bilinearity and associativity of multiplication, and entailment through vector lattice structure, can also be viewed as limitations of the model.

Bilinearity is a strong requirement to place, and has wide-ranging implications for the way meaning is represented in the model. It can be interpreted loosely as follows: components of meaning persist or diminish but do not spontaneously appear. This is particularly counter-intuitive in the case of idiom and metaphor in language. It means that, for example, both \emph{red} and \emph{herring} must contain some components relating to the meaning of \emph{red herring} which only come into play when these two words are combined in this particular order. Any other combination would give a zero product for these components. It is easy to see how this requirement arises from a context-theoretic perspective, nevertheless from a linguistic perspective it is arguably undesirable.

One potential limitation of the model is that it does not explicitly model syntax, but rather syntactic restrictions are encoded into the vector space and product itself. For example, we may assume the word \emph{square} has some component of meaning in common with the word \emph{shape}. Then we would expect this component to be preserved in the sentences \emph{He drew a square} and \emph{He drew a shape}. However, in the case of the two sentences \emph{The box is square} and \emph{*The box is shape} we would expect the second to be represented by the zero vector since it is not grammatical; \emph{square} can be a noun and an adjective, whereas \emph{shape} cannot. Distributivity of meaning means that the component of meaning that \emph{square} has in common with \emph{shape} must be disjoint with the adjectival component of the meaning of \emph{square}.

Associativity is also a very strong requirement to place; indeed \namecite{Lambek:61} introducted non-associativity into his calculus precisely to deal with examples that were not satisfactorily dealt with by his associative model \cite{Lambek:58}.

Whilst we hope that these features or boundaries are useful in their current form, it may be that with time, or for certain applications there is a reason to expand or contract certain of them, perhaps because of theoretical discoveries relating to the model of meaning as context, or for practical or linguistic reasons, if, for example, the model is found to be too restrictive to model certain linguistic phenomena.

\section{Applications to Textual Entailment}
\label{entailment}

The only existing framework for textual entailment that we are aware of is that of \namecite{Glickman:05}.\index{Glickman and Dagan!textual entailment framework} However this framework does not seem to be general enough to deal satisfactorily with many techniques used to tackle the problem since it requires interpreting the hypothesis as a logical statement.

Conversely, systems that use logical representations of language are often implemented without reference to any framework, and thus deal with the problems of representing the ambiguity and uncertainty that is inherent in handling natural language in an ad-hoc fashion.

Thus it seems what is needed is a framework which is general enough to satisfactorily incorporate purely statistical techniques and logical representations, and in addition provide guidance as to how to deal with ambiguity and uncertainty in natural language. It is this that we hope our context-theoretic framework will provide.

In this section we analyse approaches to the textual entailment problem, showing how they can be related to the context-theoretic framework, and discussing potential new approaches that are suggested by looking at them within the framework. We first discuss some simple approaches to textual entailment based on subsequence matching and measuring lexical overlap. We then look at how Glickman and Dagan's approach can be considered as a context theory in which words are represented as projections on the vector space of documents. This leads us to an implementation of our own in which we used latent Dirichlet allocation as an alternative approach to overcoming the problem of data sparseness.

\subsection{Subsequence Matching and Lexical Overlap}
\label{subsequence}
\index{subsequence matching}

We call a sequence $x \in A^*$ a ``subsequence'' of $y \in A^*$ if each element of $x$ occurs in $y$ in the same order, but with the possibility of other elements occurring in between, so for example $abba$ is a subsequence of $acabcba$ in $\{a,b,c\}^*$.
Subsequence matching compares the subsequences of two sequences: the more subsequences they have in common the more similar they are assumed to be. This idea has been used successfully in text classification \cite{Lodhi:02} and also formed the basis of the author's entry to the second Recognising Textual Entailment Challenge \cite{Clarke:06}.

If $S$ is a semigroup, $\ell^1(S)$ is a lattice ordered algebra under the multiplication of convolution:
$$(f\cdot g)(x) = \sum_{yz = x} f(y)g(z)$$
where $x,y,z \in S$, $f,g \in \ell^1(S)$.
\begin{example}[Subsequence matching]
Consider the algebra $\ell^1(A^*)$ for some alphabet $A$. This has a basis consisting of elements $e_x$ for $x \in A^*$, where $e_x$ the function that is $1$ on $x$ and $0$ elsewhere. In particular $e_\epsilon$ is a unity for the algebra.
Define $\xi(a) = \frac{1}{2}(e_a + e_\epsilon)$; then $\langle A, \ell^1(S), \xi \rangle$ is a context theory.
Under this context theory, a sequence $x$ completely entails $y$ if and only if it is a subsequence of $y$. In our experiments, we have shown that this type of context theory can perform significantly better than straightforward lexical overlap \cite{Clarke:06}. Many variations on this idea are possible, for example using more complex mappings from $A^*$ to $\ell^1(A^*)$.
\end{example}

\begin{example}[Lexical overlap]
The simplest approach to textual entailment is to measure the degree of lexical overlap:\index{lexical overlap} the proportion of words in the hypothesis sentence that are contained in the text sentence \cite{Dagan:05}.
This approach can be described as a context theory in terms of a free commutative semigroup on a set $A$, defined by $A^*/\equiv$ where $x \equiv y$ in $A^*$ if the symbols making up $x$ can be reordered to make $y$.
Then define $\xi'$ by $\xi'(a) = \frac{1}{2}(e_{[a]} + e_{[\epsilon]})$
where $[a]$ is the equivalence class of $a$ in $A^*/\equiv$. Then $\langle A, \ell^1(S/\equiv), \xi' \rangle$ is a context theory in which entailment is defined by lexical overlap. More complex definitions of $\hat{x}$ can be used, for example to weight different words by their probabilities.
\end{example}


\subsection{Document Projections}
\label{document-projections}

\noindent \namecite{Glickman:05} give a probabilistic definition of entailment in terms of ``possible worlds'' which they use to justify their lexical entailment model based on occurrences of words in web documents. They estimate the lexical entailment probability $\text{\textsc{lep}}(u,v)$ to be
$$\text{\textsc{lep}}(u,v) \simeq \frac{n_{u,v}}{n_v}$$
where $n_v$ and $n_{u,v}$ denote the number of documents that the word $v$ occurs in and the words $u$ and $v$ both occur in respectively. From the context theoretic perspective, we view the set of documents the word occurs in as its context vector. To describe this situation in terms of a context theory, consider the vector space $L^\infty(D)$ where $D$ is the set of documents. With each word $u$ we associate an operator $P_u$ on this vector space by
$$P_u e_d = \left\{\begin{array}{ll} e_d & \text{if $u$ occurs in document $d$} \\ 0 & \text{otherwise.} \end{array}\right.$$
where $e_d$ is the basis element associated with document $d \in D$. $P_u$ is a projection, that is $P_uP_u = P_u$; it projects onto the space of documents that $u$ occurs in. These projections are clearly commutative (they are in fact \emph{band projections}): $P_uP_v = P_vP_u = P_u \land P_v$ projects onto the space of documents in which both $u$ and $v$ occur.

In their paper, Glickman and Dagan assume that probabilities can be attached to individual words, as we do, although they interpret these as the probability that a word is ``true'' in a possible world. In their interpretation, a document corresponds to a possible world, and a word is true in that world if it occurs in the document.

They do not, however, determine these probabilities directly; instead they make assumptions about how the entailment probability of a sentence depends on lexical entailment probability. Although they do not state this, the reason for this is presumably data sparseness: they assume that a sentence is true if all its lexical components are true: this will only happen if all the words occur in the same document. For any sizeable sentence this is extremely unlikely, hence their alternative approach.

It is nevertheless useful to consider this idea from a context theoretic perspective. The probability of a term being true can be estimated as the proportion of documents it occurs in. This is the same as the context theoretic probability defined by the linear functional $\phi$, which we may think of as determined by a vector $p$ in $L^\infty(D)$ given by $p(d) = 1/|D|$ for all $d \in D$. In general, for an operator $U$ on $L^\infty(D)$ the context theoretic probability of $U$ is defined as
$$\phi(U) = \|U^+p\|_1 - \|U^-p\|_1,$$
where $U^+ = U \lor 0$ and $U^- = (-U) \lor 0$ and the lattice operations are defined by the Riesz-Kantorovich formula (Example \ref{riesz}).
The probability of a term is then $\phi(P_u) = n_u /|D|$. More generally, the context theoretic representation of an expression $x = u_1u_2\ldots u_m$ is $P_x = P_{u_1}P_{u_2}\ldots P_{u_m}$. This is clearly a semigroup homomorphism (the representation of $xy$ is the product of the representations of $x$ and $y$), and thus together with the linear functional $\phi$ defines a context theory\index{context theory} for the set of words.

The degree to which $x$ entails $y$ is then given by $\phi(P_x\land P_y) / \phi(P_x)$. This corresponds directly to Glickman and Dagan's entailment ``confidence''; it is simply the proportion of documents that contain all the terms of $x$ which also contain all the terms of $y$.

\subsection{Latent Dirichlet Projections}

The formulation in the previous section suggests an alternative approach to that of Glickman and Dagan to cope with the data sparseness problem. We consider the finite data available $D$ as a sample from a corpus model $D'$; the vector $p$ then becomes a probability distribution over the documents in $D'$. In our own experiments, we used latent Dirichlet allocation \cite{Blei:03} to build a corpus model based on a subset of around 380,000 documents from the Gigaword corpus. Having the corpus model allows us to consider an infinite array of possible documents, and thus we can use our context-theoretic definition of entailment since there is no problem of data sparseness.

\begin{figure*}
\begin{center}
\psset{xunit=1mm,yunit=1mm,runit=1mm}
\psset{linewidth=0.3,dotsep=1,hatchwidth=0.3,hatchsep=1.5,shadowsize=1}
\psset{dotsize=0.7 2.5,dotscale=1 1,fillcolor=black}
\psset{arrowsize=1 2,arrowlength=1,arrowinset=0.25,tbarsize=0.7 5,bracketlength=0.15,rbracketlength=0.15}
\begin{pspicture}(0,0)(135,65)
\rput{0}(8.67,22.14){\psellipse[](0,0)(8.67,-8.57)}
\rput{0}(43.33,22.14){\psellipse[](0,0)(8.67,-8.57)}
\rput{0}(78,22.14){\psellipse[](0,0)(8.67,-8.57)}
\rput{0}(112.67,22.14){\psellipse[](0,0)(8.67,-8.57)}
\rput{0}(95.33,56.43){\psellipse[](0,0)(8.67,-8.57)}
\psline[arrowsize=3 2]{->}(52,22.14)(69.33,22.14)
\psline[arrowsize=3 2]{->}(17.33,22.14)(34.67,22.14)
\psline[arrowsize=3 2]{->}(86.67,22.14)(104,22.14)
\psline[arrowsize=3 2]{->}(95.33,47.86)(112.67,30.71)
\rput(8.67,9.29){$\alpha$}
\rput(43.33,9.29){$\theta$}
\rput(78,9.29){$z$}
\rput(112.67,9.29){$w$}
\rput(82.33,56.43){$\beta$}
\pspolygon[arrowscale=2 2](60.67,39.29)(130,39.29)(130,5)(60.67,5)
\rput(65,35){$N$}
\pspolygon[arrowscale=2 2](25,45)(135,45)(135,0)(25,0)
\end{pspicture}
\caption{Graphical representation of the Dirichlet model. The inner box shows the choices that are repeated for each word in the document; the outer box the choice that is made for each document; the parameters outside the boxes are constant for the model.}
\label{graphical}
\end{center}
\end{figure*}


Latent Dirichlet allocation (LDA) follows the same vein as Latent Semantic Analysis (LSA) \cite{Deerwester:90} and Probabilistic Latent Semantic Analysis (PLSA) \cite{Hofmann:99} in that it can be used to build models of corpora in which words within a document are considered to be exchangeable; so that a document is treated as a bag of words. LSA performs a singular value decomposition on the matrix of words and documents which brings out hidden ``latent'' similarities in meaning between words, even though they may not occur together.

In contrast PLSA and LDA provide probabilistic models of corpora using Bayesian methods. LDA differs from PLSA in that, while the latter assumes a fixed number of documents, LDA assumes that the data at hand is a sample from an infinite set of documents, allowing new documents to be assigned probabilities in a straightforward manner.

\begin{figure}
\begin{center}
\framebox[7.5cm]{\parbox{7cm}{
\begin{enumerate}
\item Choose $\theta \sim$ Dirichlet$(\alpha)$
\item For each of the $N$ words:
\begin{enumerate}
\item Choose $z \sim$ Multinomial$(\theta)$
\item Choose $w$ according to $p(w|z)$
\end{enumerate}
\end{enumerate}}}
\end{center}
\caption{Generative process assumed in the Dirichlet model}
\label{generative}
\end{figure}

Figure \ref{graphical} shows a graphical representation of the latent Dirichlet allocation generative model, and figure \ref{generative} shows how the model generates a document of length $N$. In this model, the probability of occurrence of a word $w$ in a document is considered to be a multinomial variable conditioned on a $k$-dimensional ``topic'' variable $z$. The number of topics $k$ is generally chosen to be much fewer than the number of possible words, so that topics provide a ``bottleneck'' through which the latent similarity in meaning between words becomes exposed.

The topic variable is assumed to follow a multinomial distribution parameterised by a $k$-dimensional variable $\theta$, satisfying $$\sum_{i=1}^k \theta_i = 1,$$
and which is in turn assumed to follow a Dirichlet distribution. The Dirichlet distribution is itself parameterised by a $k$-dimensional vector $\alpha$. The components of this vector can be viewed as determining the marginal probabilities of topics, since:
\begin{eqnarray*}
p(z_i) & = & \int p(z_i|\theta)p(\theta)d\theta\\
	   & = & \int \theta_i p(\theta)d\theta.
\end{eqnarray*}
This is just the expected value of $\theta_i$, which is given by
$$p(z_i) = \frac{\alpha_i}{\sum_j \alpha_j}.$$

The model is thus entirely specified by $\alpha$ and the conditional probabilies $p(w|z)$ which we can assume are specified in a $k\times V$ matrix $\beta$ where $V$ is the number of words in the vocabulary. The parameters $\alpha$ and $\beta$ can be estimated from a corpus of documents by a variational expectation maximisation algorithm, as described by \namecite{Blei:03}.

Latent Dirichlet allocation was applied by \namecite{Blei:03} to the tasks of document modelling, document classification and collaborative filtering. They compare latent Dirichlet allocation to several techniques including probabilistic latent semantic analysis; latent Dirichlet allocation outperforms these on all of the applications. Recently, latent Dirichlet allocation has been applied to the task of word sense disambiguation \cite{Cai:07,Boyd-Graber:07} with significant success.

Consider the vector space $L^\infty(A^*)$ for some alphabet $A$, the space of all bounded functions on possible documents. In this approach, we define the representation of a string $x$ to be a projection $P_x$ on the subspace representing the (infinite) set of documents in which all the words in string $x$ occur. Again we define a vector $q(x)$ for $x\in A^*$ where $q(x)$ is the probability of document $x$ in the corpus model, we then define a linear functional $\phi$ for an operator $U$ on $L^\infty(A^*)$ as before by $\phi(U) = \|U^+q\|_1 - \|U^-q\|_1$. $\phi(P_x)$ is thus the probability that a document chosen at random contains all the words that occur in string $x$. In order to estimate $\phi(P_x)$ we have to integrate over the Dirichlet parameter $\theta$:
$$\phi(P_x) = \int_\theta\left(\prod_{a\in x}p_\theta(a)\right)p(\theta)d\theta,$$
where by $a\in x$ we mean that the word $a$ occurs in string $x$, and $p_\theta(a)$ is the probability of observing word $a$ in a document generated by the parameter $\theta$. We estimate this by
$$p_\theta(a) \simeq 1 - \left(1 - \sum_z p(a|z)p(z|\theta)\right)^N,$$
where we have assumed a fixed document length $N$. The above formula is an estimate of the probability of a word occurring at least once in a document of length $N$, the sum over the topic variable $z$ is the probability that the word $a$ occurs at any one point in a document given the parameter $\theta$. We approximated the integral using Monte-Carlo sampling to generate values of $\theta$ according to the Dirichlet distribution.

\begin{table}
\begin{center}
\begin{tabular}{|l||l|l|}
\hline
\textbf{Model} & \textbf{Accuracy} & \textbf{CWS}\\
\hline\hline
Dirichlet ($10^6$) & 0.584 & 0.630\\
Dirichlet ($10^7$) & 0.576 & 0.642\\
\hline
Bayer (MITRE) & 0.586 & 0.617 \\
Glickman (Bar Ilan) & 0.586 & 0.572\\
Jijkoun (Amsterdam) & 0.552 & 0.559\\
Newman (Dublin) & 0.565 & 0.6\\
\hline
\end{tabular}
\vspace{0.1cm}
\caption{Results obtained with our Latent Dirichlet projection model on the data from the first Recognising Textual Entailment Challenge for two document lengths
$N = 10^6$ and $N = 10^7$
using a cut-off for the degree of entailment of
$0.5$
at which entailment was regarded as holding.}
\label{table:lda-results}
\end{center}
\end{table}

We built a latent Dirichlet allocation model using \namecite{Blei:03}'s implementation on documents from the British National Corpus, using 100 topics. We evaluated this model on the 800 entailment pairs from the first Recognising Textual Entailment Challenge test set.\footnote{We have so far only used data from the first challenge, since we performed the experiment before the other challenges had taken place.} Results were comparable to those obtained by \namecite{Glickman:05} (see Table \ref{table:lda-results}).   In this table, Accuracy is the accuracy on the test set, consisting of 800 entailment pairs, and CWS is the confidence weighted score; see \cite{Dagan:05} for the definition. The differences between the accuracy values in the table are not statistically significant because of the small dataset, although all accuracies in the table are significantly better than chance at the 1\% level. The accuracy of the model is considerably lower than the state of the art, which is around 75\% \cite{Bar-Haim:06}. We experimented with various document lengths and found very long documents ($N = 10^6$ and $N = 10^7$) to work best.


It is important to note that because the LDA model is commutative, the resulting context algebra must also be commutative, which is clearly far from ideal in modelling natural language.

\section{The Model of Clark, Coecke and Sadrzadeh}

One of the most sophisticated proposals for a method of composition is that of \namecite{Clark:08} and the more recent implementation of \cite{Grefenstette:11}. In this section, we will show how their model can be described as a context theory.

The authors describe the syntactic element of their construction using pregroups \cite{Lambek:01}, a formalism which simplifies the syntactic calculus of \cite{Lambek:58}. These can be described in terms of \textbf{partially ordered monoids}, a monoid $G$ with a partial ordering $\le$ satisfying $x \le y$ implies $xz \le yz$ and $zx \le zy$ for all $x,y,z\in G$.

\begin{definition}[Pregroup]\index{pregroup|textbf}
Let $G$ be a partially ordered monoid. Then $G$ is called a pregroup if for each $x\in G$ there are elements $x^l$ and $x^r$ in $G$ such that
\begin{eqnarray}
x^lx \le &1\\
xx^r \le &1\\
1 \le &xx^l\\
1 \le &x^rx
\end{eqnarray}
\end{definition}
If $x,y\in G$, we call $y$ a \textbf{reduction} of $x$ if $y$ can be obtained from $x$ using only rules (1) and (2) above.

Pregroup grammars are defined by freely generating a pregroup on a set of basic grammatical types. Words are then represented as elements formed from these basic types, for example:
$$
\begin{array}{ccc}
\text{John} & \text{likes} & \text{Mary}\\
\pi & \pi^r s o^l & o
\end{array}
$$
where $\pi$, $s$ and $o$ are the basic types for first person singular, statement and object, respectively. It is easy to see that the above sentence reduces to type $s$ under the pregroup reductions.

As \namecite{Clark:08} note, their construction can be generalised by endowing the grammatical type of a word with a vector nature, in addition to its semantics. We use this slightly more general construction to allow us to formulate it in the context-theoretic framework. We define an \textbf{elementary meaning space} to be the tensor product space $V = S\otimes P$ where $S$ is a vector space representing meanings of words and $P$ is a vector space with an orthonormal basis corresponding to the basic grammatical types in a pregroup grammar and their adjoints. We assume that meanings of words live in the tensor algebra space $T(V)$, defined by
\[T(V) = \mathbb{R}\oplus V \oplus (V\otimes V) \oplus (V\otimes V\otimes V) \oplus \cdots\]
For an element $v$ in a particular tensor power of $V$, such that $v = (s_1\otimes p_1)\otimes (s_2\otimes p_2)\otimes \cdots\otimes(s_n\otimes p_n)$, where the $p_i$ are basis vectors of $P$, then we can recover a complex grammatical type for $v$ as the product $\gamma(v) = \gamma_1\gamma_2\cdots\gamma_n$, where $\gamma_i$ is the basic grammatical type corresponding to $p_i$. We will call the vectors such as this which have a single complex type (i.e.~they are not formed from a weighted sum of more than one type) \textbf{unambiguous}.

We also assume that words are represented by vectors whose grammatical type is \textbf{irreduceable}, i.e.~there is no pregroup reduction possible on the type. We define $\Gamma(T(V))$ as the vector space generated by all such vectors.

We will now define a product $\cdot$ on $\Gamma(T(V))$ that will make it an algebra. To do this, it suffices to define the product between two elements $u_1,u_2$ which are unambiguous and whose grammatical type is basic,  i.e.~they can be viewed as elements of $V$. The definition of the product on the rest of the space follows from the assumption of distributivity. We define:
$$u_1\cdot u_2 = \left\lbrace \begin{array}{ll}
u_1\otimes u_2\ \  & \text{if $\gamma(u_1\otimes u_2)$ is irreduceable}\\
\\
\langle u_1, u_2\rangle & \text{otherwise.}
\end{array} \right.$$
This product is bilinear, since for a particular pair of basis elements, only one of the above two conditions will apply, and both the tensor and inner products are bilinear functions. Moreover, it corresponds to composed and reduced word vectors, as defined in \cite{Clark:08}.

To see how this works on our example sentence above, we assume we have vectors for the meanings of the three words, which we write as $v_{\text{word}}$. We assume for the purpose of this example that the word \emph{like} is represented as a product state composed of three vectors, one for each basic grammatical type. This removes any potentially interesting semantics, but allows us to demonstrate the product in a simple manner. We write this as follows:
$$
\begin{array}{ccccc}
\text{John} & \hbox{\quad} & \text{likes} & \hbox{\quad} & \text{Mary}\\
(v_{\text{John}}\otimes e_\pi) & \cdot & (v_{\text{likes},1}\otimes e_{\pi^r})\cdot
(v_{\text{likes},2}\otimes e_s) \cdot
(v_{\text{likes},3}\otimes e_{o^l}) & \cdot & (v_{\text{Mary}}\otimes e_o)
\end{array}
$$
where $e_\gamma$ is the orthornormal basis vector corresponding to basic grammatical type $\gamma$. More interesting representations of \emph{like} would consist of sums over similar vectors. Computing this product from left to right:
$$
\begin{array}{crccl}
& (v_{\text{John}}\otimes e_\pi) \cdot
(v_{\text{likes},1}\otimes e_{\pi^r})\cdot
& (v_{\text{likes},2}\otimes e_s) & \cdot & 
(v_{\text{likes},3}\otimes e_{o^l})\cdot
(v_{\text{Mary}}\otimes e_o)\\
= &
\langle v_{\text{John}},v_{\text{likes},1}\rangle
& (v_{\text{likes},2}\otimes e_s) & \cdot &
(v_{\text{likes},3}\otimes e_{o^l}) \cdot
(v_{\text{Mary}}\otimes e_o)\\
= &
\langle v_{\text{John}},v_{\text{likes},1}\rangle
& (v_{\text{likes},2}\otimes e_s) & \otimes &
(v_{\text{likes},3}\otimes e_{o^l}) \cdot
(v_{\text{Mary}}\otimes e_o)\\
= &
\langle v_{\text{John}},v_{\text{likes},1}\rangle
\langle v_{\text{likes},3},v_{\text{Mary}}\rangle
& (v_{\text{likes},2}\otimes e_s) &
\end{array}\\$$
As we would expect in this simplified example the product is a scalar multiple of the second vector for $\emph{like}$, with the type of a statement.


This construction thus allows us to represent complex grammatical types, similar to \namecite{Clark:08}, however it also allows us to take weighted sums of these complex types, giving us a powerful method of expressing the syntactic and semantic ambiguity of lexical semantics.

%

%
%
%
%
%
%
%
%

\section{Conclusions and Future Work}

We have presented a context-theoretic framework for natural language semantics. The
framework is founded on the idea that meaning in natural language can be determined
by context, and is inspired by techniques that make use of statistical properties of
language by analysing large text corpora. Such techniques can generally be viewed as
representing language in terms of vectors. These techniques are currently used in appli-
cations such as textual entailment recognition, however the lack of a theory of meaning
that incorporates these techniques means that they are often used in a somewhat ad-
hoc manner. The purpose behind the framework is to provide a unified theoretical
foundation for such techniques so that they may used in a principled manner.

By formalising the notion of ``meaning as context'' we have been able to build a
mathematical model that informs us about the nature of meaning under this paradigm.
Specifically, it gives us a theory about how to represent words and phrases using
vectors, and tells us that the product of two meanings should be distributive and
associative. It also gives us an interpretation of the inherent lattice structure on these
vector spaces as defining the relation of entailment. It also tells us how to measure the
size of the vector representation of a string in such a way that the size corresponds to
the probability of the string.

We have demonstrated that the framework encompasses several related approaches to compositional distributional semantics, including those based on a predefined composition operation such as addition  \cite{Mitchell:08,Landauer:97,Foltz:98} or the tensor product \cite{Smolensky:90,Clark:07,Widdows:08}, matrix multiplication \cite{Rudolph:10}, and the more sophisticated construction of \namecite{Clark:08}.


\subsection{Practical Investigations}

Section \ref{entailment} raises many possibilities for the design of systems to recognise textual
entailment within the framework
\begin{itemize}
\item Variations on substring matching: experiments with different weighting
         schemes for substrings, allowing partial commutativity of words or
         phrases, and replacing words with vectors representing their context,
         using tensor products of these vectors instead of concatenation.
\item Extensions of Glickman and Dagan's approach and our own
         context-theoretic approach using latent Dirichlet allocation, perhaps using
         other corpus models based on n-grams or other models in which words
         do not commute, or a combination of context theories based on
         commutative and non-commutative models.
\item The LDA model we used is a commutative one. This is a considerable simplification of what is possible within the context-theoretic framework; it would be interesting to investigate methods of incorporating non-commutativity into the model.
\item Implementations based on the approach to representing uncertainty in logical semantics
        similar to those described in \cite{Clarke:07}.
\end{itemize}
All of these ideas could be evaluated using the data sets from the Recognising Textual
Entailment Challenges.

There are many approaches to textual entailment that we have not considered here; we conjecture that variations of many of them could be described within our framework. We leave the task of investigating the relationship between these approaches and our framework to further work.

Other areas that we are investigating, together with researchers at the University of Sussex, is the possibility of learning finite-dimensional algebras directly from corpus data, along the lines of \cite{Guevara:11} and \cite{Baroni:10}.

One question we have not addressed in this paper is the feasibility of computing with algebraic representations. Although this question is highly dependent on the particular context theory chosen, it is possible that general algorithms for computation within this framework could be found; this is another area that we intend to address in further work.

\subsection{Theoretical Investigations}

Although the context-theoretic framework is an abstraction of the model of meaning as context, it would be good to have a complete understanding of the model and the types of context theories that it allows. Tying down these properties would allow us to define algebras that could truly be called ``context theories''.

The context-theoretic framework shares a lot of properties with the study of free probability \cite{Voiculescu:97}. It would be interesting to investigate whether ideas from free probability would carry over to context-theoretic semantics.

Although we have related our model to many techniques described in the literature, we still have to investigate its relationship with other models such as that of \namecite{Song:03} and \namecite{Guevara:11}.

     We have not given much consideration here to the issue of multi-word expres-
sions and non-compositionality. What predictions does the context-theoretic framework
make about non-compositionality? Answering this may lead us to new techniques for
recognising and handling multi-word expressions and non-compositionality.

     Of course it is hard to predict the benefits that may result from what we have
presented, since we have given a way of thinking about meaning in natural language
that in many respects is new. This new way of thinking opens the door to the unification
of logic-based and vector-based methods in computational linguistics, and the potential
fruits of this union are many.

\bibliographystyle{fullname}
\bibliography{contexts}

\end{document}